\documentclass[conference]{IEEEtran}
\usepackage{cite}
\usepackage[svgnames,table]{xcolor}
\usepackage{paralist,array}
\usepackage{svg}
\usepackage{amsmath,amssymb,amsfonts}
\usepackage{algorithmic}
\usepackage{graphicx}
\usepackage{textcomp}
\usepackage{booktabs}
\usepackage{colortbl}
\usepackage{pifont}
\usepackage{float}
\usepackage[ruled,vlined]{algorithm2e}
\usepackage{enumitem}
\usepackage{multirow}
\usepackage{afterpage}
\usepackage[multiple]{footmisc}
\usepackage{makecell}
\usepackage{derivative}

\def\BibTeX{{\rm B\kern-.05em{\sc i\kern-.025em b}\kern-.08em
    T\kern-.1667em\lower.7ex\hbox{E}\kern-.125emX}}
    
\makeatletter\let\MPtrue\@minipagetrue\makeatother
    
\newcolumntype{P}[1]{>{\centering\arraybackslash}p{#1}}
\newcolumntype{M}[1]{>{\centering\arraybackslash}m{#1}}

\begin{document}


\title{\textbf{\textit{Exploring the Intersection between Neural Architecture Search and Continual Learning}}}




\makeatletter
\newcommand{\linebreakand}{%
  \end{@IEEEauthorhalign}
  \hfill\mbox{}\par
  \mbox{}\hfill\begin{@IEEEauthorhalign}
}
\makeatother








\author{

\IEEEauthorblockN{
Mohamed Shahawy\IEEEauthorrefmark{1},
Elhadj Benkhelifa\IEEEauthorrefmark{2}, and
David White\IEEEauthorrefmark{3}}

\IEEEauthorblockA{Smarts Systems, AI and Cybersecurity Research Centre (SSAICS),
Staffordshire University\\
Staffordshire, UK\\
Email: 
\IEEEauthorrefmark{1}mohamed.shahawy@research.staffs.ac.uk,
\IEEEauthorrefmark{2}e.benkhelifa@staffs.ac.uk,
\IEEEauthorrefmark{3}david.white1@staffs.ac.uk
}

}

\maketitle


\begin{abstract}

Despite the significant advances achieved in Artificial Neural Networks (ANNs), their design process remains notoriously tedious, depending primarily on intuition, experience and trial-and-error. This human-dependent process is often time-consuming and prone to errors. Furthermore, the models are generally bound to their training contexts, with no considerations to their surrounding environments. Continual adaptiveness and automation of neural networks is of paramount importance to several domains where model accessibility is limited after deployment (e.g IoT devices, self-driving vehicles, etc.). Additionally, even accessible models require frequent maintenance post-deployment to overcome issues such as Concept/Data Drift, which can be cumbersome and restrictive. By leveraging and combining approaches from Neural Architecture Search (NAS) and Continual Learning (CL), more robust and adaptive agents can be developed. This study conducts the first extensive review on the intersection between NAS and CL, formalizing the prospective Continually-Adaptive Neural Networks (CANNs) paradigm and outlining research directions for lifelong autonomous ANNs.


\end{abstract}


\section{Introduction}


Deep Learning has led to notable advances across a number of applications, specifically non-procedural tasks, such as image classiﬁcation or maneuvering autonomous vehicles. By enabling computers to accomplish these tasks at a near (or sometimes even surpassing) human-level performance \cite{otoole_2007}, tasks that previously relied on human-intervention can now become autonomous.

Despite all the incredible achievements accomplished in the field of Artificial Intelligence, there remains a number of hurdles that keep AI models from advancing further. Some Artificial Neural Networks (ANNs) have out-performed humans at a number of tasks, such as playing chess \cite{hassabis_2017}, reading comprehension \cite{darlington_2021}, audio transcription \cite{negrao_2021}, and several other areas \cite{sun_2015,yu_2016,khaw_2017,blair_2019}. However, even the most advanced models still cannot compete with a child's cognitive abilities (reasoning, abstract thinking, complex comprehension, etc.) \cite{gopnik_1999,zador_2019}. It is thought that human intelligence arises not only from learned knowledge, but also through innate mechanisms that have evolved over thousands of years \cite{seung_2013}.

Inspired by the Darwinian evolutionary theory, \textit{Neuroevolution} is an area that aims to evolves neural networks rather than develop them through classical means. This evolutionary approach attempts to mimic the brain's complex structure and capabilities by generating populations of competing individuals (neural architectures), essentially simulating the ``survival of the fittest" process. By doing so, the network topology and its weights can be encoded into the optimization process and the best fitting combination of architectures and parameters can be searched for in the solution space (as presented in the iconic work by Stanley \& Miikkulainen in 2002 \cite{stanley_2002}). 

It was later discovered, however, that multi-objective optimization of the neural architecture and its weights yielded less effective results than multi-stage single-objective optimization (i.e. optimizing the architecture and its parameters consecutively rather than concurrently) \cite{liu_enas_2021}. This AutoML area, dubbed \textit{Neural Architecture Search} (NAS), was formalized by Zoph \& Le and started trending in recent years \cite{zoph_2017}. The aim behind NAS is to leverage optimization algorithms, such as Reinforcement Learning and Evolutionary Algorithms, to provide more robust means of neural architecture design relative to the classic, manual trial-and-error approach.

Another struggle modern AI faces lies in certain domains where the input changes over the course of time. Typically, ANNs tend to partially or completely overwrite learned knowledge upon retraining for new tasks, a phenomenon referred to as \textit{Catastrophic Forgetting}, or \textit{Catastrophic Interference} \cite{mccloskey_1989,mcclelland_1995}. To address this problem, the field of \textit{Continual Learning} (CL) was introduced (sometimes referred to as \textit{Lifelong Learning} \cite{soltoggio_2018,parisi_2019,anthes_2019}, \textit{(Task-/Domain-/Class-)Incremental Learning} \cite{delange_2021,zhang_2020,luo_2020,van_2022}, and \textit{Sequential Learning} \cite{mccloskey_1989,robins_2004}).

Continual Learning aims to overcome Catastrophic Forgetting by retaining some form of implicit or explicit memory within the model, or by having continuous access to the tasks' datasets. Although complete prevention of information loss may not be attainable, the trade-off between generalizing the model to adapt to a continuous input stream and consolidating learned knowledge can be optimized. In other words, CL attempts to find a balanced solution on the stability-plasticity Pareto front, maximizing shared learning representations and adaptability to underlying non-stationary data distributions.

\subsection{Biological Inspiration} \label{subsec:biological_inspiration}

One of the reasons humans excel at continually adapting to new streams of information without forgetting previously-attained knowledge is that our experiences are generalized and largely overlap \cite{zevin_2006,chen_2016}. In addition to the human brain's innate Continual Learning abilities, numerous other adaptive qualities (synaptic plasticity, neuromodulation, neuroplasticity, etc.) have been key components in the human evolution to intelligence \cite{cleeremans_2011,bennett_2021}.

Research in the field of Neuropsychology has revealed that the evolutionary advances accomplished by our species can be attributed to two fundamental mechanisms: \textit{Associative Learning} and \textit{Memory Formation} \cite{coolidge_2020}.

\textit{Synaptic Plasticity}, or the ability to change synaptic strengths, is thought to underlie both the learning and memory mechanisms in mammalian brains \cite{takeuchi_2014,feldman_2020}. Hebb's learning rule, which was postulated by Donald Hebb in 1949, bridged the fields of psychology and neuroscience (now known as neuropsychology) by modelling associative learning on a cellular level \cite{hebb_1949}. The premise behind the rule is that the learning process in biological neural networks is activity-dependent; if a pre-synaptic neuron A is repeatedly successful in activating a post-synaptic neuron B, the strength of the synapse will gradually increase (typically referred to as \textit{Synaptic Potentiation}). Similarly, \textit{Synaptic Depression} (or the weakening of synaptic strength) occurs when a pre-synaptic neuron A consistently does not take part in firing a post-synaptic neuron B. This learning model paved the way for one of the simplest and earliest learning rules in Artificial Neural Networks.

The mechanism underlying memory formation and retention is considered one of the most compelling and challenging research areas in the field of neuroscience \cite{lynch_2004,dringenberg_2020}. Intuitively, it is thought that a particularly strong connections between a set of neurons that resulted from high-frequency, persistent strengthening of synapses is a direct representation of memory on a cellular level. On the other hand, lack of access to those neurons would result in a gradual decay of unused memories over time \cite{tsumoto_1993,escobar_2007}. It is worth noting that researches on cognitive psychology have revealed that memory entails a learning process in itself \cite{tulving_1995}, and thus the mechanisms for both processes are interleaved.

The Complementary Learning Systems framework, which was articulated by McClelland et al. in 1995, defines the distinct roles of the mammalian hippocampal and neocortical systems on learning and episodic memory \cite{mcclelland_1995}. The theory holds that the hippocampus is responsible for rapid learning and retaining short-term, sparse memories, which can later be replayed to the Prefrontal Cortex for more efficient encoding and long-term storage. Although the CLS is characterized as a theory, it has had over 25 years' worth of empirical data generally supporting its premise \cite{oreilly_2011,blakeman_2020}.

The CLS model has inspired several Continual Learning approaches for ANNs, often with explicit short-/long-term memories (modelling the Hippocampal Complex/Medial Prefrontal Cortex) and a replay mechanism through a virtual sleep phase \cite{kemker_fearnet_2018,shin_2018}. Although many other biological neural functions have not been modelled in Artificial Neural Networks yet, there has been a growing interest in fields pertaining to plasticity in computational models \cite{soltoggio_2018}.

\begin{figure*}[!ht]
\centering
\includegraphics[width=0.98\textwidth]{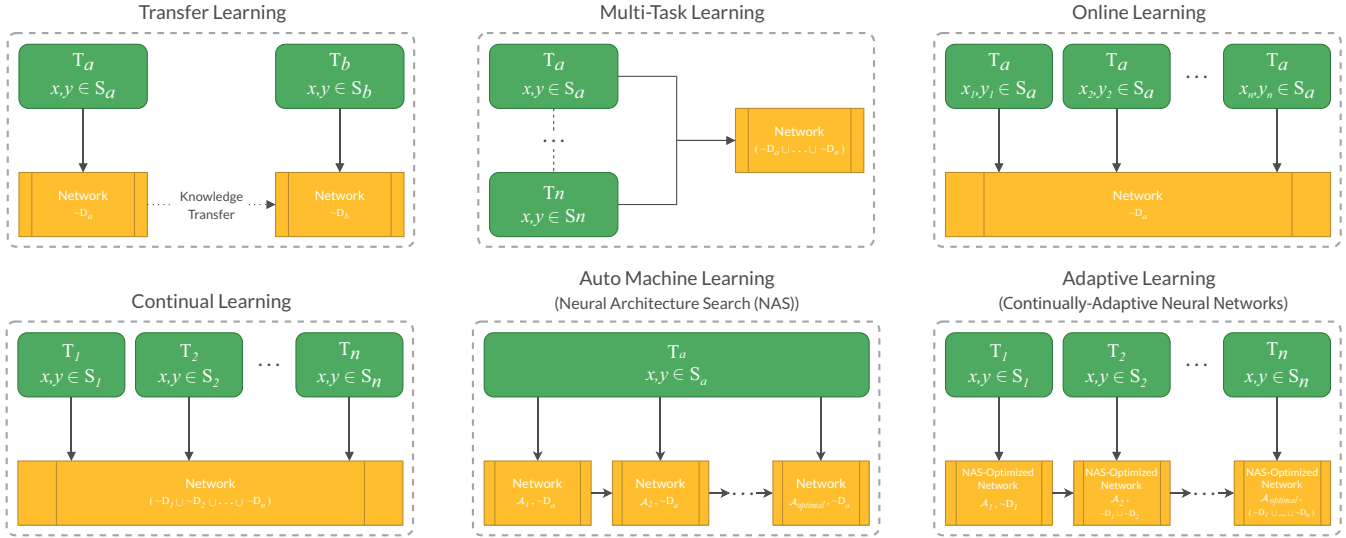}
\caption{Comparative illustrations for Adaptive Learning and its neighboring Machine Learning paradigms (inspired by \cite{delange_2021})}
\label{fig:learning_paradigms}
\vspace{-1.0em}
\end{figure*}

\subsection{Motivation, Scope, and Contributions to Knowledge}

Continually adaptive features in ANNs are becoming more imminent than ever; several fields that have limited access to their deployed models are bound to their fixed, pre-deployment states and cannot improve their functionality to fit their changing surroundings. These fields include autonomous vehicles \cite{pierre_2018,sun_2018,wehbe_2019,nose_2019,si_2019,ferdaus_2019,shaheen_2021}, robotics \cite{thrun_1995,lesort_2020,chen_2020,zheng_2021,xie_2021,kahn_2021}, Edge and IoT devices \cite{cheng_2012,song_2018,bao_2019,chathoth_2021}, and numerous others \cite{rasouli_2015,patra_2020,elskhawy_2020,chauhan_2020,srivastava_2021,liu_2021,park_2021}.

In addition to the importance of Continual Learning in modern AI applications, the classical process of developing neural networks is heavily-reliant on manual human input; from defining a network architecture to tweaking hyperparameters, several parameters are dependent on experts' knowledge as well as trial-and-error \cite{ren_2022}. The AutoML field was introduced to facilitate more robust Machine Learning solutions for non-experts by pipelining automated optimization techniques, such as \textit{Neural Architecture Search} (NAS) and Hyperparameter Optimization. This area, however, is still in its infancy and few studies have attempted to achieve an end-to-end self-developing ANN \cite{raghavan_2019,kaplunovich_2021}.

Intuitively, the more statically-developed a neural network is (e.g. fixed network topology, fixed training learning representation, etc.), the less sensitive it will be to its surrounding environment. By minimizing or completely eliminating manual input from the development of ANNs, a network can become significantly less biased and thus more adaptive to fit the tasks' contexts. This concept of self-managing autonomous agents has been explored in the literature, however it is usually discussed within bounded domains (namely Neuroevolution and Unsupervised Learning) \cite{mouret_2014,soltoggio_2018,miconi_2018}. 

\textbf{Aims and Contributions}. To expand the currently narrow and specialized mindset associated with ANNs, we extend and generalize the inspiring ideas previously proposed beyond the scope of unsupervised learning and Evolutionary Algorithms. \textit{Continually-Adaptive Neural Networks} (CANNs) aim to ultimately eliminate human-intervention from the development and maintenance of ANNs by combining Neural Architecture Search and Continual Learning solutions. Functional to that aim, this study:

\begin{enumerate}
    \item Reviews Continual Learning approaches and proposes an improved categorization scheme for memory models
    \item Reviews relevant works in the Neural Architecture Search field
    \item Explores the intersection between the two domains, formalizes the automated lifelong supervised learning paradigm, and briefly reviews additional data automation techniques (hyperparameter optimization and autonomous data collection/pre-processing/feature engineering)
    \item Identifies the gaps and potential future directions for NAS, CL, and CANNs
\end{enumerate}

Although there are several reviews conducted on Continual Learning \cite{parisi_2019,delange_2021}, AutoML \cite{nagarajah_2019,waring_2020,he_2021}, and NAS \cite{elsken_2019,jaafra_2019,wistuba_2019,ren_2022}, the intersection between NAS and Continual Learning has not been formalized or reviewed despite the significant number of existing models in that area. To capture the scope of this study, Fig. \ref{fig:learning_paradigms} visualizes the learning representation of Continual Learning (task-incremental setting), Neural Architecture Search, their intersection (CANNs), as well as neighboring paradigms to emphasize their distinction.


\textbf{Paper Outline}. To investigate hybrid Continual Learning and Neural Architecture Search models, we first conduct two distinct reviews on the latest developments in each area (Sections \ref{sec:cl} \& \ref{sec:nas}). After which, the intersection of the two fields (dubbed the Continually-Adaptive Neural Networks paradigm) is formalized, outlining ideal characteristics to achieve autonomous and adaptive agents. Lastly, Section \ref{sec:future_directions} summarizes the limitations of the existing methodologies and delineates potential future research directions.

\section{Continual Learning Approaches} \label{sec:cl}

Continual Learning aims to maximize the capacity for simultaneous solutions by generalizing previous experiences to coincide with new tasks (a feature known as \textit{Forward Transfer} (FWT) of knowledge), whilst also consolidating and maintaining previous knowledge (referred to as \textit{Backward Transfer} (BWT)). Balancing this trade-off between stability and plasticity is essential for a sustainable CL model, and is commonly known as the \textit{Stability-Plasticity dilemma} \cite{grossberg_1982,mermillod_2013}. For instance, consider two incremental tasks, $\mathcal{T}_i$ and $\mathcal{T}_j$, where $\mathcal{T}_i \prec \mathcal{T}_j$.

Forward Transfer is the impact of $\mathcal{T}_i$ on the performance of $\mathcal{T}_j$; if a preceding task is advantageous to the learning process of a future task, it is considered positive FWT, which potentially leads to \textit{``zero-shot"/``few-shot" learning} \cite{lopez-paz_2017}. Negative FWT, on the other hand, occurs when a previous task hinders the learning process of a future task, essentially making a model less pliable to fit new tasks.

Backward Transfer is the effect the learning process of task $\mathcal{T}_j$ has on the performance of task $\mathcal{T}_i$. There exists positive BWT if learning $\mathcal{T}_j$ increases the performance of the preceding task $\mathcal{T}_i$. Conversely, negative BWT exists when learning $\mathcal{T}_j$ decreases the performance on $\mathcal{T}_i$ (i.e Catastrophic Forgetting) \cite{lopez-paz_2017, parisi_2019}.

Since any model is constrained by a finite memory capacity and the number of tasks is potentially unbounded, forgetting can be inevitable. As the number of tasks increases, convergence becomes more compromised, leading to a decline in the model performance. Therefore, the goal behind CL models is not to prevent Catastrophic Forgetting, but to alleviate it and selectively minimize complete loss of knowledge, a feature referred to as \textit{Graceful Forgetting} \cite{aljundi_2018,aljundi_2019,delange_2021,ling_2021}. 

\textbf{Types of Continual Learning}. Given the incremental nature of Continual Learning, there are three fundamental settings that may be encountered \cite{van_2022}. For a set of incremental tasks $\mathcal{T}$, each task $t$ can have $d \geq 1$ incremental datasets $\mathcal{D}^{t}_{d}=\{\mathcal{X}_{d}, \mathcal{Y}_{d}\}\:\; s.t.\:\; t\in\mathcal{T}$, where $\mathcal{P}(\mathcal{X}), \mathcal{P}(\mathcal{Y})$ are the probability distributions of the training data and output labels, respectively.

\begin{enumerate}[label=(\roman*)]
    \item \textit{Domain-Incremental Learning} (DIL): A single task with dynamic input distribution and a static label space. $|\mathcal{T}| = 1, \mathcal{P}(\mathcal{X}_i)\neq\mathcal{P}(\mathcal{X}_j), \mathcal{P}(\mathcal{Y}_i)=\mathcal{P}(\mathcal{Y}_j), \forall (i\neq j)\in\mathcal{D}$ 
    
    \item \textit{Class-Incremental Learning} (CIL): A single task with dynamic input distribution and an intersecting (but growing) label space. CIL is typically modelled with an expanding single-head output layer. $|\mathcal{T}| = 1, \mathcal{P}(\mathcal{X}_i)\neq\mathcal{P}(\mathcal{X}_j), \mathcal{Y}_i\cap\mathcal{Y}_j, \forall (i\neq j)\in\mathcal{D}$ 
    
    \item \textit{Task-Incremental Learning} (TIL): A set of distinct tasks with disjoint input and output spaces. TIL is typically modelled with a multi-head output layer to accommodate the dissimilar probability distributions of the tasks. $|\mathcal{T}| > 1, \mathcal{P}(\mathcal{X}_i)\neq\mathcal{P}(\mathcal{X}_j), \mathcal{P}(\mathcal{Y}_i)\neq\mathcal{P}(\mathcal{Y}_j), \forall (i\neq j)\in\mathcal{D}$ 
\end{enumerate}

In addition to the incremental settings, there are several different approaches to retain memory in CL models. While there have already been a few proposed taxonomies \cite{farquhar_2018,aljundi_2019,delange_2021}, we propose a categorization scheme (Fig. \ref{fig:cl_approaches}) that adjusts the view of memory-retention models in the PANNs context.

\begin{figure}[htb]
\centering
\includegraphics[width=0.48\textwidth]{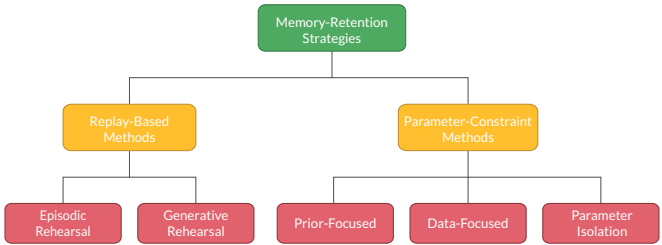}
\caption{Continual Learning memory-retention strategies}
\label{fig:cl_approaches}
\vspace{-1.25em}
\end{figure}

Previous works regarded ``static/dynamic architectures" as a family of memory models, whereas the adaptiveness of a network topology is tangent to overcoming catastrophic forgetting. For example, some replay-based approaches have fixed topologies \cite{rebuffi_2017}, while others have dynamic topologies \cite{yoon_2017}. This showcases that structural plasticity is a catalyst for memory improvement that can be combined with any strategy in Fig. \ref{fig:cl_approaches}, rather than a holistic approach for retaining knowledge. Hence, we propose to classify structural plasticity as an independent quality that, albeit relevant to memory-retention, should not be categorized as a distinct family of CL models.

\subsection{Replay-Based Methods}

The first of the two families of methods is \textit{Replay-Based} Continual Learning. Replay-Based models retain an episodic memory by storing representations that best describe the distribution of the data used for each task. These representations could simply be a raw portion of the datasets (\textit{Episodic Rehearsal}), or it could be the model itself that was trained on a previous task used to infer and generate the samples used (essentially reverse-engineering the model) (\textit{Generative Rehearsal}, often called \textit{pseudo-rehearsal}). Each sub-type of the Replay-based models has its own strengths and weaknesses:

\begin{itemize}
  \item Episodic Rehearsal: store a subset of the datasets used to later "rehearse" them in subsequent training of newer tasks. While this method might be the easiest to implement out of the 2 sub-types, the model can easily overfit to the subsets due to the limited number of samples, and does not scale well in terms of memory cost.
  \item Generative Rehearsal: store the previous tasks' distributions instead of the raw samples, which can later be used to project and generate auxiliary data for replay purposes. Although this method has not been used frequently in recent studies on CL, the recent emergence of generative networks opens a potential for new Generative Rehearsal models \cite{delange_2021}.
\end{itemize}

Both Replay-Based approaches tend to be computationally- and/or memory-expensive, especially as the number of tasks increases \cite{rebuffi_2017,parisi_2019,rolnick_2019,delange_2021}. However, by caching portions of the datasets or the models used for each task, the retraining process can support prioritization for particular tasks, thereby facilitating more control over Graceful Forgetting.

In general, Replay-Based methods draw inspiration from the biological CLS framework, where short-term memory is quickly stored to the hippocampus, to be later learned slowly and in a more generalized fashion by the neocortex (likely during sleep). For a more comprehensive review on the association between biological mechanisms and Replay-Based CL models, the recent work by Hayes et al. covers the topic extensively and in-depth \cite{hayes_2021}.

\subsection{Parameter-Constraint Methods}

The latter family of approaches is \textit{Parameter-Constraint} Continual Learning. By constraining parameters through isolation, regularization, or any form of transfer learning/fine-tuning, Parameter-Constraint models can efficiently prevent Catastrophic Forgetting whilst not wasting memory on prior data storage. Sub-types of Parameter-Constraint models are split three-fold:

\begin{itemize}
  \item Data-Focused: constrain parameters through a combination of Fine-Tuning and \textit{Knowledge Distillation} \cite{hinton_2015}, where a temporary (large and highly regularized) model is trained on a single task and is then used to transfer the knowledge to the main (smaller) network.
  \item Prior-Focused: constrain parameters through regularization. By penalizing changes to the most influential parameters in the network, Catastrophic Forgetting can be minimized. Geometrically, the model attempts to find the area of best fit (where most solutions coincide) in the solution space.
  \item Parameter-Isolation: constrain parameters through isolation. Unlike Prior-Focused models, this method completely isolates some of the weights ($\lambda = 0$) instead of just applying penalties, hence dedicating portions of the network to individual tasks.
\end{itemize}

A minor limitation with penalizing changes to the weights or completely isolating them is that the priority will typically be given to the earlier tasks, rendering Graceful Forgetting less attainable using traditional methods \cite{rusu_2016,aljundi_2017}.


By combining Replay-Based and Parameter-Constraint methods, a third, implicit family of approaches can be defined. The first method to include regularization techniques on top of an episodic memory is Gradient Episodic Memory for Continual Learning (GEM) \cite{lopez-paz_2017}. GEM aims to constrain the parameter changes pertaining to new tasks by placing an upper-bound limit on the losses of each task; the loss of every task can decrease but cannot increase.

\begin{table*}[!htp]
    \centering
    \rowcolors{0}{}{gray!5}
    \begin{tabular}{P{4.5cm} P{2cm} P{2cm} P{7.5cm}}
        \toprule
        Model & Incremental Setting & Memory Strategy & Limitations \\
        \midrule
        
        Learning Without Forgetting (LwF) \cite{li_2018} & Task-Incremental & Data-Focused
         & \begin{minipage} [t] {0.4\textwidth} \begin{itemize}[nosep,after=\strut]
            \item Highly dependent on the relevance of the tasks
            \item Training time gradually increases with the number of tasks
            \item Architecture grows linearly with each task-increment
        \end{itemize}
        \end{minipage}\\

        Incremental Classifier and Representation Learning (iCaRL) \cite{rebuffi_2017} & Class-Incremental & Episodic Rehearsal & 
        \begin{minipage} [t] {0.4\textwidth}  \begin{itemize}[nosep,after=\strut]
            \item Requires explicit storage for older tasks
            \item Can easily overfit on the tasks' exemplars subset
            \item The exemplars chosen may not be representative of the entire dataset
        \end{itemize}
        \end{minipage}\\

        Elastic Weight Consolidation (EWC) \cite{kirkpatrick_2017} & Task-Incremental & Prior-Focused &
        \begin{minipage} [t] {0.4\textwidth} \begin{itemize}[nosep,after=\strut]
            \item Strictly requires a fixed network architecture (known in advance)
            \item Numerically unstable for a large number of tasks
        \end{itemize}
        \end{minipage}\\

        Gradient Episodic Memory for Continual Learning (GEM) \cite{lopez-paz_2017} & Task-Incremental & Episodic Rehearsal \& Prior-Focused & 
        \begin{minipage} [t] {0.4\textwidth} \begin{itemize}[nosep,after=\strut]
            \item Requires explicit storage for older tasks
            \item Highly dependent on the relevance of the tasks
            \item No memory-management policy
        \end{itemize}
        \end{minipage}\\

        Memory-Aware Synapses (MAS) \cite{aljundi_2018} & Task-Incremental & Prior-Focused & \begin{minipage} [t] {0.4\textwidth} \begin{itemize}[nosep,after=\strut]
            \item Strictly requires a fixed network architecture (known in advance)
        \end{itemize}
        \end{minipage}\\

        Progressive Neural Networks (PNN) \cite{rusu_2016} & Task-Incremental & Parameter-Isolation &
        \begin{minipage} [t] {0.4\textwidth} \begin{itemize}[nosep,after=\strut]
            \item Does not facilitate positive BWT
        \end{itemize}
        \end{minipage}\\

        Experience Replay for Continual Learning (CLEAR) \cite{rolnick_2019} & Task-Incremental & Episodic Rehearsal & 
        \begin{minipage} [t] {0.4\textwidth} \begin{itemize}[nosep,after=\strut]
            \item Built on the assumption that input data stream is always iid
        \end{itemize}
        \end{minipage}\\

        Continual Learning through Synaptic Intelligence (SI) \cite{zenke_2017} & Task-Incremental & Prior-Focused & 
        \begin{minipage} [t] {0.4\textwidth} \begin{itemize}[nosep,after=\strut]
            \item Inefficient when applied to a pretrained network
            \item Tasks' order affects performance
            \item The importance of the weights is not properly estimated in some cases (as noted by the authors)
        \end{itemize}
        \end{minipage}\\

        Deep Generative Dual Memory Network for Continual Learning (DGR) \cite{kamra_2017} & Task-Incremental & Generative Rehearsal &
        \begin{minipage} [t] {0.4\textwidth} \begin{itemize}[nosep,after=\strut]
            \item Strictly requires a fixed network architecture (known in advance)
            \item Model saturates with tasks over time (performance declines as the number of tasks increases)
        \end{itemize}
        \end{minipage}\\

        Incremental Moment Matching (IMM) \cite{lee_2017} & Class-Incremental \& Domain-Incremental & Prior-Focused & 
        \begin{minipage} [t] {0.4\textwidth} \begin{itemize}[nosep,after=\strut]
            \item Requires explicit storage for previous tasks' models
        \end{itemize}
        \end{minipage}\\


        Less-Forgetting Learning (LFL) \cite{jung_2016} & Domain-Incremental & Data-Focused & 
        \begin{minipage} [t] {0.4\textwidth} \begin{itemize}[nosep,after=\strut]
            \item Highly dependent on the relevance of the tasks
            \item Susceptible to data-distribution (domain) shift \cite{aljundi_selfless_2019}
        \end{itemize}
        \end{minipage}\\

        Expert Gate \cite{aljundi_2017} & Task-Incremental & Parameter-Isolation &
        \begin{minipage} [t] {0.4\textwidth} \begin{itemize}[nosep,after=\strut]
            \item Does not facilitate positive BWT due to gating mechanism
        \end{itemize}
        \end{minipage}\\
        
        \bottomrule
    \end{tabular}
    \bigskip
    \label{tab:cl_models}
    \caption{Brief analysis of Continual Learning models from an adaptiveness perspective}
    \vspace{-1.5em}
\end{table*}


\subsection{Evaluation of Continual Learning Performance}

Despite the large body of research on Continual Learning, the benchmark tests and evaluation metrics for Continual Learning models are largely heterogeneous and very few works have been devoted to formalize an evaluation framework that captures the models' performance in dynamic environments \cite{rusu_2016,parisi_2019,lesort_2020,new_2022}. The majority of the early works relied on a rather uninformative evaluation approach, measuring only the average accuracy across all tasks. While accuracy is generally considered a reliable measure of the overall model performance, Continual Learning should be ranked based on more robust criteria that are specific to dynamic scenarios.

Lopez-Paz \& Ranzato proposed one of the first formal evaluation methods for Continual Learning models \cite{lopez-paz_2017}. Their methodology revolves around more "human-like" test environments, where the models' evaluation should be based on a large number of tasks, each with a relatively small number of examples, and the reported metrics should include not only accuracy, but also a measure of forward/backward transfer. Given that some models can yield the same accuracies, FWT and BWT can provide a better insight on the model's long-term performance. Following the same direction, Díaz-Rodríguez et al. later proposed a similar evaluation framework, with the addition of 3 metrics: model-size efficiency, samples storage-size efficiency, and computational efficiency \cite{diaz-rodriguez_2018}. While these metrics are not directly related to the tasks' performance (i.e accuracy, stability, plasticity, etc.), they are vital indicators to a model's scalability.

While the measure of FWT and BWT can provide extensive information on the behavior of a model, they do not directly estimate how well a model retains memory. Kemker et al. focused on assessing a model's ability to overcome Catastrophic Forgetting particularly \cite{kemker_2018}. The study establishes 3 (normalized AUC-ROC curve) metrics that jointly measure memory retention, the model's ability to acquire new knowledge, and the model's overall performance: $\Omega_{base}$, $\Omega_{new}$, and $\Omega_{all}$. The base metric evaluates how well the first/base task is retained after subsequent tasks have been learned. $\Omega_{new}$ measures the performance of a new task immediately after it is learned, which shows the model's few-shot learning ability. Lastly, $\Omega_{all}$ evaluates the model's overall performance in terms of retention and acquisition of knowledge.

Inspired by the aforementioned approach, Hayes et al. proposed a similar evaluation framework that is split into 3 experimental paradigms \cite{hayes_2018}. The first benchmark test simply measures the model's ability to quickly learn new tasks through continual iid data. While real-world scenarios tend to comprise of non-iid streams of data, this test is used to establish a baseline metric that can be compared with offline models. The second test evaluates the model's ability to incrementally learn new classes whilst measuring the tasks' test accuracies with each increment, which can also indicate when Catastrophic Forgetting occurs. Finally, the last test consists of a stream of organized non-iid data, resembling a realistic stream of stimuli that could be experienced by a deployed model. This framework also uses the $\Omega_{all}$ metric to measure stability, plasticity, and overall performance.

It is important to highlight that Continual Learning environments can vary to great extents. A robust evaluation framework should accommodate these variations and yield consistent metrics regardless of the application's environment. Farquhar \& Gal introduced the first desiderata for CL ranking, highlighting the desirable features in an evaluation framework \cite{farquhar_2018}. The proposed desiderata suggests that data from later tasks should resemble earlier tasks to some extent, arguing that some evaluation approaches use permuted datasets (permuted MNIST in particular \cite{kemker_2018,hayes_2018}), which is an unnecessarily excessive test. The study also claims that the output vector used in the training phase should be the same as the one used during testing. Another desirable feature in the evaluation of a CL model is to restrict retraining for older tasks; the model should not be allowed unconstrained access to early tasks' examples. Last but not least, the test should include more than 2 tasks, as success on such a small sample size cannot yield reliable performance metrics that can be applicable to larger scales.

The most recent formal study on Continual Learning evaluation metrics was proposed by New et al. in 2022 as part of the DARPA L2M research program \cite{new_2022}. The primary aim behind this evaluation framework is to robustly rank models in a domain-agnostic fashion, doing so based on 5 metrics: Performance Maintenance (across all tasks), FWT, BWT, Performance Relative to a Single Task Expert, and Sample Efficiency (ability to improve the learning process over time). Together, these metrics capture the effectiveness of memory-retention, knowledge transfer \& adaptation, and model sustainability. Not only do such metrics assist researchers in identifying the suitable models for each application domain, they also empirically consolidate all the ideal qualities for CL models proposed in previous studies. Aiming for higher ranks on this domain-agnostic adaptiveness scale further paves the way towards autonomous agents.




\section{Neural Architecture Search Approaches} \label{sec:nas}

Most popular Deep Neural Networks are designed by human experts, where the potentially hundreds of layers are manually specified \cite{szegedy_2015,simonyan_2015,he_2016,iandola_2016}. The architecture design process usually depends on the experts' educated intuition and a tedious trial-and-error process. As the tasks become more complex, the networks are normally expected to be deeper and thus their architectures become trickier to design and optimize.

In addition to the extensive knowledge required to design neural architectures, humans tend to have a fixed thinking paradigm that limits the exploration of the search space and leads to unvaried solutions. These drawbacks of manual Deep Neural Networks' design were further recognized when an automatically-tuned neural network (Auto-Net 1.0) won against human experts' designs in the ChaLearn AutoML challenge in 2015 \cite{guyon_2015,mendoza_2016,mendoza_2019}.

The field of \textit{Neural Architecture Search} (NAS) emerged to minimize or completely eliminate the limitations posed by the human factor in neural architecture-design whilst providing more robust means of finding optimal topologies. The NAS process comprises of defining 3 main components (as depicted in Fig. \ref{fig:nas_process}): a \textit{Search Space}, a \textit{Search Algorithm}, and an \textit{Evaluation Strategy} \cite{elsken_2019,ren_2022}.

\begin{figure}[ht]
\centering
\includegraphics[width=0.48\textwidth]{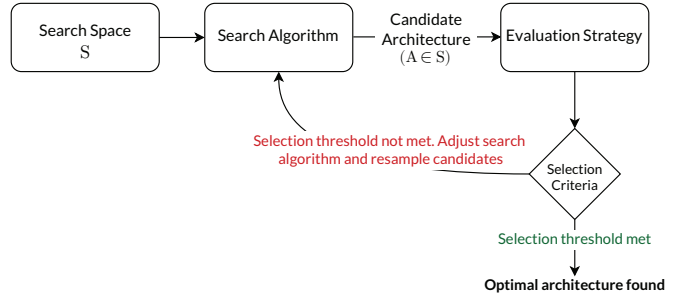}
\caption{The general Neural Architecture Search framework}
\label{fig:nas_process}
\vspace{-1em}
\end{figure}

\subsection{Search Spaces}

The first step in a typical NAS approach is defining the search space. The search space dictates the potential solutions' boundary and can have a detrimental impact on the outcome of the entire search process \cite{yu_2019}. An over-simplified search space will likely lead to shallow networks that are only capable of solving simple tasks, while a non-continuous and high dimensional space can yield more complex networks, but is difficult to optimize and is typically computationally expensive.

\subsubsection{Layer-Wise Space}

One of the simplest search spaces is the Layer-Wise space (sometimes referred to as the Stage-Wise or Chain-Structured search space \cite{wu_2019,hu_2019,ren_2022,baymurzina_2022}), as seen in the iconic early works of Zoph \& Le and Baker et al. \cite{zoph_2017,baker_2016,elsken_2019}. As implied by the name, the Layer-Wise space is used to sample a sequentially-built network from a space of layers. For a simple Layer-Wise architecture $\mathcal{A}$ consisting of $n$ layers, the $i^{th}$ layer $L_{i}$ would receive its input from $L_{i-1}$, and its output would feed into $L_{i+1}$, forming the chain $\mathcal{A} = L_{n-1} \circ \dots \circ L_{1} \circ L_{0}$ (as illustrated in Fig. \ref{fig:chain_structured_nas}).

\vspace{0.1cm}
\begin{figure}[ht]
\centering
\includegraphics[width=0.48\textwidth]{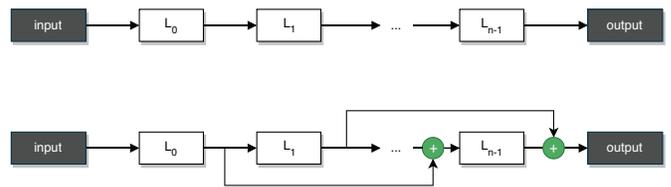}
\caption{Layer-Wise Search Space for Neural Architecture Search; A single-branch architecture sample (top) and a multi-branch counterpart (bottom)}
\label{fig:chain_structured_nas}
\end{figure}

A Layer-Wise search space can be defined through 3 parameters: The number of layers $n$, the type of layers (e.g. convolution, pooling, flattening, the number of units (in a fully-connected network), etc.), and the hyperparameters associated with the previously-defined layer types. Some studies, however, have more recently started introducing more complex design elements, such as Skip-Connections (depicted in Fig. \ref{fig:chain_structured_nas}; bottom) \cite{baker_2016,zoph_2017,cai_2018,zoph_2018,elsken_2019,gong_2019}. The addition of Skip-Connections allows the search algorithms to build multi-branch networks (similar to ResNet \cite{he_2016} and DenseNet \cite{huang_2017}) which significantly expands the search space. This expansive nature, however, is inherently greedy and resource-intensive.  


\subsubsection{Cell-Based Space}

To circumvent the high performance requirements posed by the Layer-Wise space, more recent works took advantage of the repeated fixed sub-structures in neural networks, creating cells (i.e groups of operations; see Fig. \ref{fig:cell_based_nas}) that can be reused as individual units. This modular approach to neural architecture design was manually implemented in some popular networks, such as ResNet and InceptionNet \cite{szegedy_2015,he_2016}, and is also one of the most common search spaces in NAS models. The Cell-based Space (often referred to as the Modular Search Space and the Block-Based Space) drastically reduces the resource cost by compounding the building-blocks that can be used \cite{liu_pnas_2018,zoph_2018,zhong_2018,elsken_2019,ren_2022}.

\vspace{0.05cm}
\begin{figure}[ht]
\centering
\includegraphics[width=0.48\textwidth]{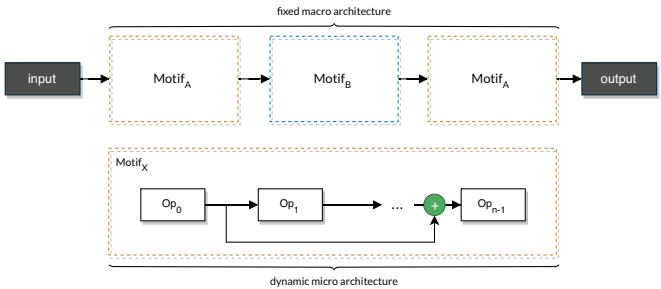}
\caption{Generic Cell-based Neural Architecture Search topology sample}
\label{fig:cell_based_nas}
\end{figure}
\vspace{-0.25em}

Zoph et al. introduced one of the top-performing popular NAS frameworks, Google's NASNet, which was developed using a Cell-Based NAS space \cite{zoph_2018}. The NASNet Search Space comprises of only 2 types of cells: a \textit{Normal Cell}, and a \textit{Reduction Cell}. The Normal Cell performs feature extraction whilst preserving the input dimensionality, whereas the Reduction Cell down-samples the given feature map to half its size. By manually defining a fixed \textit{macro-architecture} (the structure comprising of Normal/Reduction cells), optimal operations for each cell-type can be searched for. By adopting this higher level design paradigm and shifting the manual design focus from an elementary-unit level to the macro-architecture, several NAS frameworks have attained exceptional performance results \cite{zoph_2018,zhong_2018,cai_2018,cai_proxylessnas_2018,elsken_2018,real_2019}.

In addition to the reasonable resource usage the Cell-based Space offers, the modular nature of the space enables the transferability of cells back and forth between networks (i.e cell-based Transfer Learning). For instance, individual cells from NASNet were optimized on CIFAR-10, then transferred and used on ImageNet, breaking highest-accuracy records on both benchmarks (97.60\% and 82.70\%, respectively) \cite{zoph_2018}. 

Relative to the Layer-Wise space, this approach breaks down the optimization phase into smaller problems that can be computed faster. However, this reduced search-cost is leveraged at the expense of the agility of the framework. The Cell-based space requires a priori knowledge to manually pre-define the static macro-structure that constrains the search space, as opposed to the more expansive Layer-Wise space that can dynamically adapt based on the tasks' context.

\subsubsection{Hierarchical Space}

Another search space that also focuses on macro-architecture optimization is the Hierarchical Space, first introduced by Liu et al. \cite{liu_2017}. Similar to the aforementioned modular approach, the Hierarchical Space's concept is to construct sub-graphs of a topology, then use them to recursively compose an architecture. The key difference between the Cell-Based and Hierarchical spaces is that the latter uses recursive composition to build the topology rather than modular stacking.

At the core, the Hierarchical Space uses primitive operations (e.g. 1x1 convolution, 3x3 maxpooling, ReLU, etc.) to set up the first level, $\ell_{1}$. The next level $\ell_{2}$ would consist of sub-graphs, each composed of multiple primitive elements from $\ell_{1}$, and so on for each subsequent level. 

Furthermore, the composition process yields a new operation in the form of a Directed Acyclic Graph (DAG), which are typically encoded as upper-triangular adjacency matrices. More formally, each level $\ell$ in the Hierarchical Space comprises of $n$ adjacency matrices $\mathcal{O}^{(\ell)} = \langle o^{(\ell)}_{1}, o^{(\ell)}_{2}, ... ,o^{(\ell)}_{n} \rangle$, such that:


\begin{equation}
o^{(\ell)}_{i} = compose(\mathcal{O}^{(\ell-1)}) \: , \; \forall \ell = 2 , ... , L
\label{eq:hierarchical_space}
\end{equation}

Where the $compose(\cdot)$ function constructs an operation using a combination of multiple motifs from the given set of operations (the previous level). The Hierarchical Space has proven to be an efficient search space, even when optimized by naïve Search Algorithms, such as Random Search and Grid Search \cite{liu_2017,elsken_2019}.

\subsubsection{Other Search Spaces}

Earlier works on neural architecture optimization, particularly using Neuroevolution and other Metaheuristic algorithms, searched for optimal Perceptrons on a synaptic/neural level (to be referred to as the \textit{Neuronal Search Space}) \cite{stanley_2002,stanley_2009,wu_2020}. This granular approach, however, is generally limited to shallower and simpler topologies relative to modern NAS spaces \cite{liu_enas_2021}. 

Although the Layer-Wise, Cell-Based, and Hierarchical search spaces are considered some of the most commonly used spaces in the literature, there has been a surge of alternative methodologies recently. For instance, rather than manually designing a static search space, Ci et al. proposed a novel approach to evolve progressively-growing search spaces, which they call \textit{Neural Search-space Evolution} (NSE) \cite{ci_2021}. The authors of NSE claim that traditional search space-design approaches cannot effectively handle large spaces, stating that using existing state-of-the-art algorithms could take up to 27,000 (Nvidia GTX 1080Ti) GPU hours for 100 epochs' worth of search through a large, 27-operations space. However, using NSENet, that number diminishes to 4,000 GPU hours, with slightly better Top-1 accuracy score.

Due to the NAS Search Spaces' discrete nature, Gradient Optimization was inherently not considered a valid optimization approach. However, Liu et al. introduced the concept of \textit{relaxing} the structural parameters of the network, transforming the discrete search space into a continuous and differentiable space (often denoted as the \textit{Continuous Search Space}) \cite{liu_2018}. This relaxation process consists of transforming all categorical operations into a continuous choice of probabilities. A particular operation $\bar{o}_{(i,j)}$ is relaxed using softmax over the set of all operations in the space $\mathcal{O}$ as follows:

\begin{equation}
\bar{o}_{(i,j)}(x) = \sum_{o \in \mathcal{O}} \frac{e^{\alpha_{o}(i,j)}}{\sum_{o' \in \mathcal{O}}e^{\alpha_{o}(i,j)}} o(x^{(i)})
\label{eq:DARTS}
\end{equation}

Another unique continuous Search Space-design approach is \textit{Memory-Bank Representation} (MBR), introduced by Brock et al. \cite{brock_2018}. This method uses an auxiliary fully-convolutional ``HyperNet" to produce weights for randomly sampled architectures (the memory banks). In addition to dramatically reducing the candidates' training time through the HyperNet's One Shot evaluation (expanded upon in subsection \ref{subsubsec:os_eval}), this proxy representation enables approximate differentiation, which supports architectural gradient descent.

\subsection{Search Algorithms}

After defining a Search Space, the next step in a standard NAS pipeline is to create and optimize a structure using a Search Algorithm. A Search Algorithm is the approach used to sample the candidate network(s) from a given solution space to be evaluated thereafter and further optimized if needed. Most NAS Search Algorithms are similar to approaches used for hyperparameter optimization as both areas are inherently similar \cite{liashchynskyi_2019}.

\subsubsection{Random/Grid Search}

Although widely considered a naïve optimization approach due to its exhaustive nature, Random Search (RS) has proven to be useful for hyperparameter tuning \cite{bergstra_2012,mantovani_2015}. RS algorithms for NAS work by selecting a candidate network at random (using a given distribution, e.g.: uniform, normal, poisson, etc.) from the search space \cite{li_2020}. The glaring downside of such an approach is that the selection process is statistically independent and cannot be optimized given the feedback from the Evaluation Strategy. However, the lack of complexity leads to a much faster selection process and therefore great exploration capabilities, which can yield high-performing results \cite{liu_2017,yu_2019}.

Another popular exhaustive search technique used for Neural Architecture Search is Grid Search (GS). By defining the number of layers and a set of operations per layer \newline (i.e $L_{i} = \{conv3x3,maxpool3x3,dense256\}$), GS can select network architectures by trying different combinations of options from the given sets of operations. As with Random Search, the values are independent of each other in Grid Search approaches, meaning the algorithm can be parallelized. However, out of the two exhaustive algorithms, Random Search often outperforms Grid Search \cite{liashchynskyi_2019}.

\subsubsection{Reinforcement Learning}

Reinforcement Learning (RL) problems generally comprise of (i) an Agent that uses an optimization algorithm to take certain actions and achieve a given task, (ii) a Policy that defines the set of possible actions, (iii) an Environment (sometimes called an Action/Observation Space) that the Agent can interact with, and finally (iv) a Reward Function that incentivizes the Agent every iteration to reach an optimal solution. When applied to NAS, these components remain the same; the architecture generator/controller (the Agent; typically an RNN controller model) uses a set of possible operations/layers (the Policy) in a given Search Space (the Environment), which can then be assessed using an Evaluation Strategy (the Reward Function). 

NAS is widely considered to have gained popularity after the groundbreaking work by Zoph \& Le in 2017, which was one of the earliest models to achieve impressive performance on the CIFAR-10 and the Penn-Treebank datasets using an automatically-searched neural architecture \cite{zoph_2017}. Their model was based on a Reinforcement Learning search algorithm that used a Recurrent Neural Network (RNN) with the REINFORCE Policy Gradient algorithm \cite{williams_1992} as an Agent/controller that samples a Convolutional Neural Network (CNN) from a given set of CNN operations (convolutional layers with a variable number of kernels and kernel sizes, ReLU, Batch Normalization, and Skip-Connections). Some of the generated models using this approach achieved an error-rate of 3.65\% on the CIFAR-10 image dataset, outperforming ResNet and some variations of DenseNet. The highly criticized aspect of this approach is its vast computational requirements, using 800 (Nvidia K40) GPUs for 4 weeks (i.e. over half a million GPU hours in total) \cite{zoph_2017,zoph_2018}.


Google Brain's famous NASNet by Zoph et al. was also built using an RL Search Algorithm \cite{zoph_2018}. NASNet mainly differs from the initial NAS model by Zoph \& Le in that the controller uses Proximal Policy Optimization \cite{schulman_2017} instead of REINFORCE Policy Gradient, and the Search Space is Cell-Based rather than Layer-Wise. More impressively, NASNet obtained an even better error-rate (as low as 2.40\%) on the same CIFAR-10 benchmark, with the copious amount of computational requirements.

\subsubsection{Neuroevolution/Evolvable-NAS}

Two of the most prominent Search Algorithms for NAS are Neuroevolution (NE) and Evolvable Neural Architecture Search (ENAS); while both methodologies are based on Evolutionary Algorithms (EAs), NE aims to optimize both the structure and the weights of a network, whereas ENAS is only concerned with the architecture \cite{liu_enas_2021}. NE and ENAS draw inspiration from biological evolution, where an architecture corresponds to an individual in a population that could breed (other architectures with similar features) if it is well-performing, or get removed from the population otherwise. Simulating this ``survival of the fittest" approach in-silico has continued to show promising results for over 3 decades \cite{miller_1989,stanley_2009,real_2017,real_2019}.



\begin{algorithm}[ht]
\SetAlgoLined
\caption{A generic Evolvable NAS pseudocode (adapted from \cite{xie_2017})}
\SetKw{Input}{input}
\SetKw{Initialization}{initialization}
\SetKw{Output}{Output}

\SetKw{Selection}{Selection}
\SetKw{Crossover}{Crossover}
\SetKw{Mutation}{Mutation}
\SetKw{FitnessEval}{Fitness evaluation}

\Input{: A dataset D, number of generations T, number of individuals in a generation N, and the probabilities of mutation and crossover $P_m$ and $P_c$} 
\BlankLine
\Initialization{: Generate a set of randomized individuals (neural architectures) $\{\mathbb{M}_{0}, n\}_{n = 1}^{N}$ and evaluate their fitness}
\BlankLine
\For{t = 1, 2, ..., T}{
    \Selection{: select a new generation $\{\mathbb{M}^{'}_{t,n}\}_{n=1}^{N}$ based on the fitness scores of $\{\mathbb{M}_{t-1,n}\}_{n=1}^{N}$}
    \BlankLine
    \Crossover{: for each pair $[\mathbb{M}_{t,2n-1} , \mathbb{M}_{t,2n}]^{[N/2]}_{n=1}$, perform crossover with a probability of $P_c$}
    \BlankLine
    \Mutation{: for each non-crossover topologies $\{\mathbb{M}_{t,n}\}_{n=1}^{N}$, perform mutation with a probability of $P_m$}
    \BlankLine
    \FitnessEval{: compute the fitness for all generated neural architectures $\{\mathbb{M}_{t,n}\}^{N}_{n=1}$}
    
}
\BlankLine
\Output{the last generation of ANN architectures $[\mathbb{M}_{T,n}]^N_{n=1}$ with their fitness scores (typically accuracy / error)}
\end{algorithm}

Evolutionary NAS, although a relatively newer area than Neuroevolution, has gained increasing momentum since 2017 \cite{real_2017,liu_2017,xie_2017,suganuma_2017,elsken_2018,real_2019}. ENAS models typically use GAs to evolve and optimize the neural structure, then train the network using Stochastic Gradient Descent (SGD) \cite{liu_enas_2021}. AmoebaNet, proposed by Real et al. in 2019 \cite{real_2019}, applies the \textit{Tournament Selection} \cite{goldberg_1991} method, where the best candidate is selected out of a random sample set $S$ and the candidate's mutated offspring gets put back into the population (sample size $|S|=1$ would turn this into a Random Search algorithm). AmoebaNet also implements a selection approach called \textit{aging evolution} that favors younger genotypes, which puts the exploration of the search space ahead of the exploitation (i.e prevents the search algorithm from narrowing down on networks that perform well early). Tournament Selection has also been used in a number of other successful ENAS models \cite{real_2017,liu_2017,chen_2019}. On the other hand, Elsken et al. employed a multi-objective Pareto front optimization approach to sample parents inversely proportional to their density \cite{elsken_2018}.

Although RL Search Algorithms tend to be the focus of attention in the NAS research community, a survey conducted by Real et al. on RL, NE/ENAS, and RS has concluded that Reinforcement Learning and evolutionary approaches perform similarly in terms of final test accuracies. Moreover, Neuroevolution and Evolutionary NAS had a slightly better anytime-performance and were more inclined to find smaller, more efficient models \cite{real_2019}. These findings, however, do not factor in the optimization convergence profiles. Conversely to RL and EAs, Gradient Optimization (GO) excels at fast convergence over smaller periods of search time; gradient calculations provide a robust convergence trajectory every iteration whereas discrete optimization requires prolonged training to provide a sufficiently reliable approximation. 

\subsubsection{Gradient Optimization}

Following the concepts introduced by Shin et al. \cite{shin_2018} and Liu et al. \cite{liu_2018}, transforming a search space into a continuous space was popularized and thus Gradient Optimization became a viable algorithm for NAS. Widely considered the basis of differentiable NAS, Differentiable Architecture Search (DARTS) uses a continuous Cell-Based Space search space, where each cell represents a node in a sequentially-built DAG, which can then be simultaneously relaxed and optimized using GO \cite{liu_2018}. The differentiable nature of the Continuous Search Space also enables DARTS to jointly optimize the architecture and the weights of the network using gradient descent.


On the other hand, Dong \& Yang proposed to forgo the bi-level optimization used in DARTS and instead prioritized reducing the search cost through sequential 2-stage optimization \cite{dong_gdas_2019}. In other words, while DARTS optimizes the weights and topology in the form of a nested loop (which unnecessarily repeats the relaxation process of the search space DAG), GDAS performs each optimization process separately in a consecutive manner, which reduces the complexity of the search. Additionally, GDAS incrementally relaxes each edge of the search space graph rather than the entire structure simultaneously, further breaking down the optimization process into smaller and more manageable problems. Consequently, this approach significantly reduces the overall search efficiency and is able to find competitively-performing architectures in just 4 GPU hours, making it one of the most efficient NAS methods in the state-of-the-art. 



\subsubsection{Other Search Algorithms}

In addition to the aforementioned mainstream optimization algorithms for NAS, Surrogate Model-Based Optimization (SMBO) (often interchangeable with Bayesian Optimization (BO) \cite{bergstra_2011,liu_pnas_2018}) has been consistently successful in many areas for almost a decade \cite{bergstra_2013,shahriari_2015,mendoza_2016,zhou_2019,white_2019}. Based on Bayes Theorem, BO works by building a probabilistic model (called the \textit{surrogate model}) of the objective/cost function. A \textit{selection/acquisition function} can then be used to sample data points from the posterior, which would iteratively approximate the solution surface. There are numerous surrogate model and acquisition function definitions; however, most BO models use Gaussian Processes (GPs) as surrogate models and Expected Improvement Criterion (modeled below) as selection functions \cite{bergstra_2011}.

\begin{equation}
E.I_{y^{*}}(x) = \int_{-\infty}^{\infty} max(y - y^{*},0) \, p(y|x) \,dy
\label{eq:expected_improvement_formula}
\end{equation}

the Expected Improvement (EI) function (Eq. \ref{eq:expected_improvement_formula}) is built over the surrogate model $p(y|x)$, where $y$ is the objective function, $y^{*}$ is the maximum observed value of $y$, and $x$ is the data-point (the hyperparameter in the case of hyperparameter optimization, or the network in case of NAS). The integral (or, in practice, a discrete and bounded variation of which) yields the approximate solution space surface, where the maximum observed value $y^{*}$ represents the global maximum, or the optimal solution.

Closely related to NE/ENAS, Swarm Intelligence (SI) optimization approaches also have been applied to NAS and yielded impressive results. Niu et al. proposed a Particle Swarm Optimization (PSO)-based NAS model that has achieved state-of-the-art accuracy (95.38\% on CIFAR10) in just 0.2 GPU days whilst maintaining an incredibly small model size \cite{niu_2019}. Other SI algorithms, such as Ant Colony Optimization \cite{byla_2019,lankford_2020,suda_2020}, have also been briefly explored in the NAS context. Further research might be needed, however, to cover the unexplored SI approaches (Microbial Intelligence, Fish School Search, Glowworm Algorithm, etc.) for NAS.


\subsection{Evaluation Strategies}



\begin{table*}[!htb]
    \centering
    \rowcolors{0}{}{gray!5}
    \begin{tabular}{M{1.5cm} M{3cm} M{1cm} M{2cm} M{2cm} M{2cm} M{1.5cm} M{1.5cm}}
        \toprule
        Search Algorithm & Reference / Model & Year & Search Space & Evaluation Strategy & GPU Usage (days)\footnotemark & Params (millions) & CIFAR10 Error (\%) \\
        \midrule
        
        
        \cellcolor{white} 
        & Luo et al. (NAONet Random-WS) \cite{luo_2018} & 2018 & Hierarchical & OS & 0.25 & 3.90 & 3.92 \\
        
        \cellcolor{white} 
        & Brock et al. (SMASHv2) \cite{brock_2018} & 2018 & MBR & OS & 16 & 4.60 & 4.03 \\
        
        \cellcolor{white}
        & Li \& Talwalkar (RandomNAS) \cite{li_2020} & 2020 & Cell-Based & OS & 2.7 & 4.30 & 2.85 \\ 
        
        \cellcolor{white}
        \multirow{-7}{*}{RS} & Zhang et al. (Random-NSAS) \cite{zhang_rs_2020} & 2020 & Cell-Based & OS & 0.7 & 3.08 & 2.64 \\

        \midrule

        \cellcolor{white}
        & Liu et al. (DARTS (second order)) \cite{liu_2018} & 2018 & Continuous Cell-Based & OS & 4 & 3.30 & 2.76 \\
        
        \cellcolor{white}
        & Cai et al. (ProxylessNAS-G) \cite{cai_proxylessnas_2018} & 2018 & Continuous Cell-Based & OS & 8.33 & 5.70 & 2.06 \\

        \cellcolor{white}
        & Dong \& Yang (GDAS) \cite{dong_gdas_2019} & 2019 & Continuous Cell-Based & OS & 0.17 & 2.50 & 2.82 \\
        
        \cellcolor{white}
        \multirow{-6.25}{*}{GO} & Zhou et al. (BayesNAS) \cite{zhou_2019} & 2019 & Continuous Hierarchical & OS & 0.1 & 3.40 & 2.41 \\

        \midrule

        \cellcolor{white}
        & Liu et al. (PNAS) \cite{liu_pnas_2018} & 2018 & Cell-Based & LFE & - & 3.20 & 3.41 \\
        
        \cellcolor{white}
        \multirow{-2}{*}{SMBO} & Kandasmy et al. (NASBOT) \cite{kandasamy_2018} & 2018 & Layer-Wise & LFE & 1.6 & - & 12.09 \\

        \midrule

        \cellcolor{white}
        & Baker et al. (MetaQNN) \cite{baker_2016} & 2016 & Layer-Wise & Full Training & 90 $\pm$ 10 & 11.20 & 6.92 \\

        \cellcolor{white} 
        & Zoph \& Le (NAS) \cite{zoph_2017} & 2017 & Layer-Wise & Full Training & 22,400 & 37.40 & 3.65 \\

        \cellcolor{white}
        & Zoph et al. (NASNet-A) \cite{zoph_2018} & 2018 & Cell-Based & Full Training & 2,000 & 3.30 & 3.41 \\
        
        \cellcolor{white}
        & Cai et al. (EAS) \cite{cai_2018} & 2018 & Hierarchical & Weight Inheritance & 17.5 $\pm$ 7.5 & 23.40 & 4.23 \\
        
        \cellcolor{white}
        & Pham et al. (Efficient NAS) \cite{pham_2018} & 2018 & Cell-Based & OS & 0.45 & 4.60 & 2.89 \\
        
        \cellcolor{white}
        \multirow{-9.5}{*}{RL} & Ding et al. (BNAS) \cite{ding_2021} & 2021 & Cell-Based & OS & 0.19 & 4.80 & 2.88 \\

        \midrule

        \cellcolor{white}
        & Stanley \& Miikkulainen (NEAT) \cite{stanley_2002} & 2002 & Neuronal & - & - & - & - \\
        
        \cellcolor{white}
        & Stanley et al. (HyperNEAT) \cite{stanley_2009} & 2009 & Neuronal & - & - & - & - \\
        
        \cellcolor{white}
        & Real et al. (Large-Scale Evolution) \cite{real_2017} & 2017 & Layer-Wise & Weight Inheritance & 2,600 \cite{ren_2022} & 5.40 & 5.40 \\
        
        \cellcolor{white}
        & Liu et al. (Hierarchical EAS-Evolution) \cite{liu_2017} & 2018 & Hierarchical & Full Training & 300 & 15.70 & 3.75 \\
        
        \cellcolor{white}
        & Real et al. (AmoebaNET-A) \cite{real_2019} & 2019 & Cell-Based & Full Training & 3,150 & 3.20 & 3.34 \\
        
        \cellcolor{white}
        & Elsken et al. (LEMONADE) \cite{elsken_2018} & 2019 & Cell-Based & Weight Inheritance & 40 $\pm$ 16 & 3.40 & 3.60 \\
        
        \cellcolor{white}
        & Lu et al. (NSGA-Net) \cite{lu_2019} & 2019 & Cell-Based & LFE & 4 & 3.30 & 2.75 \\
        
        \cellcolor{white}
        & Guo et al. (SPOS) \cite{guo_2020} & 2020 & Cell-Based & OS & - & - & - \\
        
        \cellcolor{white}
        \multirow{-15}{*}{NE/ENAS} & Yang et al. (CARS) \cite{yang_2020} & 2020 & Cell-Based & OS & 0.4 & 3.60 & 2.62 \\

        \midrule
        
        \cellcolor{white}
        & Niu et al. (PNAS) \cite{niu_2019} & 2019 & Cell-Based & Weight Inheritance & 0.2 & 2.44 & 4.62 \\
        
        \cellcolor{white}
        & Byla \& Pang (DeepSwarm) \cite{byla_2019} & 2019 & Layer-Wise & LFE & - & - & 11.31 \\

        \cellcolor{white}
        & Mo et al. (SA-NAS-c) \cite{mo_2021} & 2021 & Cell-Based & LFE & 0.1 & 3.20 & 2.53 \\
        
        \cellcolor{white}
        \multirow{-6.5}{*}{MO} & Shahawy \& Benkhelifa (HiveNAS) \cite{shahawy_2022} & 2022 & Layer-Wise & LFE & 0.3 & 1.39 & 8.90 \\

        \bottomrule
    \end{tabular}
    \bigskip
    \caption{Performance comparison for NAS models based on their Search Space, Search Algorithm, and Evaluation Strategy}
    \label{tab:nas_models}
    \vspace{-1.5em}
\end{table*}

\afterpage{\footnotetext{Although regularly used as a performance metric in the literature, GPU models vary across the experiments conducted. These numbers serve only as an approximate estimation of computational-demands.}}

In order to obtain feedback to guide the Search Algorithm and further optimize the sampled solutions, an Evaluation Strategy is need to estimate the samples' performance. The choice of Evaluation Strategy does not tremendously affect the search process, rather it can have a great impact on the computational demands of the model as the evaluation process is performed at every iteration \cite{hu_2020}. 

\subsubsection{Full Training}

Considered the simplest form of performance estimation, \textit{Full Training} (often termed  ``Training from Scratch") simply trains every candidate network until convergence and then estimates the models' accuracy and/or other metrics. This approach, although straightforward and provides the most accurate feedback, is the most computationally demanding Evaluation Strategy; every candidate network has to undergo a full training and testing phase to produce a single feedback data sample.

Models using this naïve approach of evaluation normally require a large number of GPUs running in parallel, leading to excessively high resource demands (in the order of thousands of GPU days) \cite{zoph_2017,zoph_2018,real_2019,kyriakides_2020,chen_enas_2020}. These unfeasible requirements prompted the development of new evaluation methodologies that reduce the performance estimation time.

\subsubsection{Lower Fidelity Estimation}

One way to effectively reduce the evaluation time is by training each candidate architecture on (i) just a subset of the data \cite{klein_2018,liu_2019}, (ii) for a fewer number of epochs (early-stop) \cite{zoph_2018}, (iii) using a down-scaled version of the dataset (lower resolution images, etc.) \cite{gong_2019}, and/or (iv) using a compressed version of the candidate model \cite{zoph_2018,real_2019,xu_2021}. This method is commonly referred to as \textit{Lower Fidelity Estimation} (LFE; sometimes denoted as Proxy Task Metrics).

Lower Fidelity Estimation, although significantly faster and cheaper than the traditional Full Training approach, has its downsides as well. Compressing the evaluation training set/model presents an obvious risk of bias in the candidates' performance estimation process \cite{zela_2018}. The resulting bias, however, can be negligible given that the Search Algorithms can typically be guided using the relative performance rankings of the candidate networks. Nevertheless, if the difference between the biased and fully-trained metrics is too large, the candidates' rankings might become relatively inconclusive and a multi-fidelity approach might be needed \cite{trofimov_2020,yang_2021,li_2017}.

\subsubsection{Weight Inheritance}

Another way to circumvent high computational demands is to rely on \textit{Weight Inheritance} (WI) from a parent model (often denoted as \textit{Network Morphism}), an approach similar to Net2Net's knowledge transfer \cite{chen_2015}. By initializing each candidate network to a relevant function, or by simply using the Full Training approach once and then only fine-tuning the weights every subsequent generation, the evaluation time can be cut down significantly.

Furthermore, because most Search Algorithms (specifically ENAS-based models \cite{liu_enas_2021}) do not tremendously disrupt the architectures of new individuals, Weight Inheritance can be applied to the unchanged areas of the topology and most of the parameters can be swiftly passed down across generations without requiring retraining from scratch. This method has been proven to efficiently evaluate candidate networks by numerous NAS models \cite{elsken_2018,cai_2018,cai_proxylessnas_2018,weng_2019}.

\subsubsection{Performance Predictors}

An alternative Evaluation Strategy is through the use of \textit{Performance Predictors}. A particularly popular performance prediction approach is \textit{Learning Curve Extrapolation} (LCE), which initializes the candidate networks through partial training (early-stopping) and then uses a surrogate model to estimate the generalization of the final learning curve \cite{baker_2017}. Some studies consider Learning Curve Extrapolation to be a subset of Lower Fidelity Estimation given its early-stopping nature \cite{white_2021}; however, an argument could be posed that the use of fewer epochs to train candidate networks is only one step in this Evaluation Strategy rather than its whole premise, hence Learning Curve Extrapolation is typically considered its own distinct methodology \cite{elsken_2019,ru_2021}.

While many models implementing Learning Curve Extrapolation have achieved state-of-the-art performance with impressively low computational requirements \cite{domhan_2015,klein_2016,baker_2017}, the surrogate model often requires extensive training (both weight and hyperparameter optimization) to yield satisfactory extrapolation performance. Nevertheless, Ru et al. have shown that sufficient surrogate model performance can be achieved under a \textit{fixed-hyperparameter} framework \cite{ru_2021}.

\subsubsection{One-Shot Models} \label{subsubsec:os_eval}

A particularly unique approach to search for neural architectures is One-Shot (OS) models, which heavily rely on \textit{weight-sharing} (not to be confused with the One-Shot Learning paradigm that aims to train a model using one input sample). The premise of OS methods is to build a single over-parameterized network (a large DAG, often defined as the \textit{super-network}/\textit{supernet}) whose sub-graphs are individual candidate models, sharing their weights on common edges. The training process for one super-network is only slightly more complex than that of any individual candidate in the search space \cite{wistuba_2019}. This approach is often categorized as a distinct ``Search Strategy" rather than any individual component from Fig. \ref{fig:nas_process} \cite{bender_2018}. However, given that the primary aim of OS NAS is to reduce search cost through a single training phase, we classify the One-Shot NAS approach as an Evaluation Strategy to conform with the taxonomy adopted by the research community.

Brock et al. proposed one of the earliest OS models, SMASH \cite{brock_2018}. SMASH operates by generating pseudo-random weights through a surrogate ``HyperNet", which (although randomly-generated) are sufficiently optimal to provide a relative measure for the candidate sub-graphs. Sampling hundreds of architectures can be done in under 1 GPU day due to the concurrent architecture/weights search, which (compared to other Search Algorithms/Evaluation Strategies) is significantly fast.


In the context of adaptive learning, one apparent limitation of OS models is that the super-network is static and thus constrains the Search Space to its sub-graphs. Furthermore, One-Shot models are (i) immensely restricted by the GPU's capacity given that the supernet normally needs to be loaded into the memory during the search process, and (ii) they inherently do not enable individual sub-graph optimization due to weight-sharing.

Notwithstanding the achievements accomplished in the NAS domain, frameworks generally assume a non-incremental experimental setup where all data is present prior to architecture generation, which limits their use to static environments. In the next section, we explore how NAS can be used in lifelong learning settings and formalize autonomous adaptiveness for ANNs.

\section{Continually-Adaptive Neural Networks} \label{sec:panns}

Humans have evolved to be capable of general learning and skill-refinement; the essence of our intelligence is captured by our ability to continuously assimilate new knowledge and adapt to the varying stimuli we are exposed to (i.e Classical Conditioning) \cite{thomas_1980,stern_2017}. Most modern ANNs are models of what is referred to as \textit{dedicated intelligence} \cite{boyd_2011}, or domain-specific intelligence, which is a product of the particularly restrictive (albeit efficient) single-task methodologies commonly used today \cite{soltoggio_2018}.

Human-like intelligence has long been the ultimate goal of research in AI \cite{baum_2010}. Although the definition of ``human intelligence" is largely tacit, one of its most defining features is \textit{adaptive self-modification} \cite{goertzel_2014}. 

\subsection{Definitions} \label{subsec:pann_definitions}

In the context of Artificial Neural Networks, the existing definitions of the term ``adaptive" are inconsistent. Most early studies described weight-modification as a type of adaptiveness (implying that all neural networks are fundamentally adaptive) \cite{widrow_1990,palnitkar_2004}, while more recent works generally confined the definition to specific areas (i.e. domain-adaptive \cite{ghifary_2014}, mode-adaptive \cite{zhang_2018}, etc.). 

Derived from Hebb's principle on Synaptic Plasticity, the term ``Plastic" Artificial Neural Networks (PANNs) has been used to describe adaptiveness on a cellular/synaptic level \cite{stanley_phd_2004}. These self-managed autonomous agents, however, are exclusively researched in Unsupervised Learning settings using Neuroevolution \cite{mouret_2014,soltoggio_2018}. To break out of the restricted scope of Neuroevolution and address continual autonomy in Supervised Learning environments, we propose and formalize the \textit{Continually-Adaptive Neural Networks} (CANNs) paradigm. We define an ANN as \textit{continually-adaptive} if it is:


\vspace{0.25cm}
\begin{enumerate}

\item \label{item:criterion_1} \textbf{fully-autonomous} throughout its life-cycle,
\item \label{item:criterion_2} \textbf{inherently continual} (i.e. has CL attributes, regardless of its Incremental Learning setting), and
\item \label{item:criterion_3} capable of handling and adapting to an \textbf{infinite stream of incremental input} (with allowed degradation of performance over time; see Section \ref{subsec:desiderata}-\ref{item:graceful_forgetting}).

\end{enumerate}
\vspace{0.25cm}


\begin{figure}[!ht]
\centering
\includegraphics[width=0.48\textwidth]{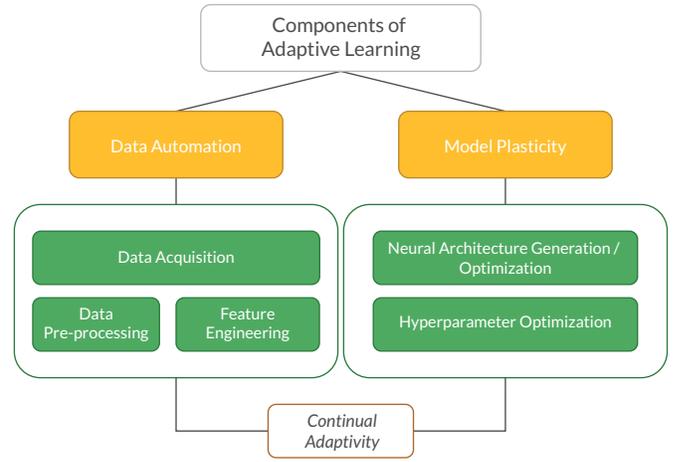}
\caption{Proposed taxonomy for Continually-Adaptive Neural Networks}
\label{fig:main_taxonomy}
\vspace{-1em}
\end{figure}

Although there are numerous additional components that could be considered adaptive (e.g. \textit{learning rule plasticity}), Fig. \ref{fig:main_taxonomy} defines the baseline, mandatory elements that would ensure end-to-end adaptiveness in a typical supervised learning ANN life-cycle.


Moreover, it is important to highlight that -- according to Criterion 2 -- a CANN should not simply be an iterative AutoML pipeline. While there exists several approaches that can automate pre-training data preparation and model generation, it is desirable for every component to have Continual Learning attributes; the learning process should not be iterated with statistical independence for every task encountered, but rather enable Forward and Backward Transfer wherever possible.

\subsection{CANN Components}

A CANN framework comprises of 2 essential components: \textit{Data Automation} and \textit{Model Plasticity}.

The first stage in a standard Machine Learning pipeline is Data Preparation, which includes all procedures required before the training stage. These procedures include \textit{Data Acquisition}, \textit{Data Pre-processing}, and \textit{Feature Engineering} (as shown in Fig. \ref{fig:data_automation}) \cite{waring_2020,ashmore_2021,he_2021}. 

\subsubsection{Data Acquisition}

In particular real world applications, a continuous source of data (e.g. a sensor, live camera, etc.) is attached to the model, rendering the Data Acquisition step unnecessary. However, in other scenarios, the data-source is either insufficient (provides a limited amount of data) or simply does not exist. Considering the importance of the dataset's size and its impact on a model's performance, a number of approaches have emerged to extend limited data sources. The most common techniques to expand small data in an automated setting are \textit{Web Search / Scrape} and \textit{Data Synthesis}.

With the increasing availability of public datasets, web-scraping \cite{munzert_2014,kunang_2018,diouf_2019} and dedicated APIs\footnote{htttps://www.datasearch.elsevier.com}\footnote{https://datasetsearch.research.google.com/} \cite{dojchinovski_2018} have become viable solutions for automated Data Acquisition. The main concerns with such Data Search approaches are: 

\begin{itemize}
    \item The legality and ethics of web-scraping \cite{krotov_2018}.
    \item Search results are likely to contain irrelevant data.
    \item Unpredictable data quality; without a supplementary data-quality verification technique, the acquired data is unreliable.
\end{itemize}

Data Synthesis, in contrast, offers a more controllable Data Acquisition environment that overcomes all the listed drawbacks posed by Web Search/Scrape. The fast-growing research on generative models and data simulators has shown that synthetic data can be an impressive alternative when real data is unattainable. The mesmerizing realism produced by \textit{Generative Adversarial Networks} (GANs) \cite{goodfellow_2014} also proves that synthetic data can be almost indistinguishable from their real counterpart whilst providing granular control of the results \cite{bowles_2018}. Moreover, tools like OpenAI Gym, Unreal Engine, and even the Grand Theft Auto game have been successfully used to simulate entire scenes for image classification and segmentation \cite{wu_2018,kar_2019}. The apparent limitation of Data Synthesis is that the process (in an automated context) requires an initial dataset to establish a baseline distribution for the synthesizer to match.

\begin{figure}[!t]
\centering
\includegraphics[width=0.48\textwidth]{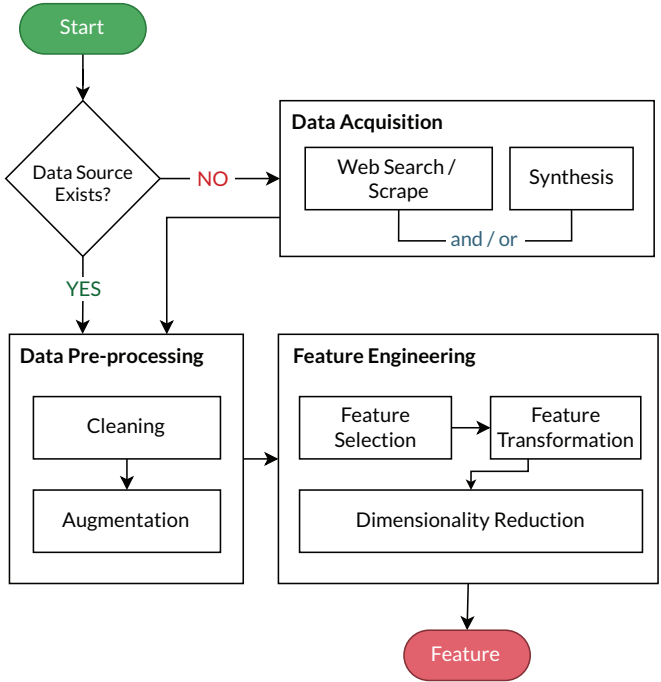}
\caption{Continual Data Automation flowchart}
\label{fig:data_automation}
\vspace{-1.25em}
\end{figure}

\subsubsection{Data Pre-processing}

Whether the data is directly fed to the framework through a continuous source or externally acquired, Data Cleaning is inevitable. The cleaning process includes removing noise/outliers, handling missing data (commonly denoted as \textit{NaN}s), and eliminating inconsistencies \cite{li_2021}. Although the Data Pre-processing stage is heavily dependent on the domain of the problem (image, text, audio signals, etc.), some data cleaning algorithms, such as BoostClean \cite{krishnan_2017} and AlphaClean \cite{krishnan_2019}, are domain-agnostic and therefore enable generic automation. Moreover, Continuous Data Cleaning frameworks that leverage the user's feedback on cleaning preferences have been briefly explored \cite{volkovs_2014}.

Amongst the issues that commonly arise during Data Preparation are class imbalance (uneven data per label), and lack of variability/over-fitting. By augmenting the data (i.e using affine transformations on image datasets or pitch-scaling on audio signals; we refer the reader to the survey by He et al. \cite{he_2021} for a Data Augmentation taxonomy), new data points can be generated to help balance the dataset and increase the generalization ability of the model. A byproduct of Data Augmentation is dataset extension and so it can also be regarded as a Data Acquisition approach. However, the consensus is that augmentation is primarily used for its regularization qualities and is thus considered a part of the Data Pre-Processing stage \cite{shorten_2019,he_2021}.

\subsubsection{Feature Engineering}

\begin{table*}[!htb]
    \centering
    \rowcolors{0}{}{gray!5}
    \begin{tabular}{M{3cm} M{1.15cm} M{1.35cm} M{1.4cm} M{1.75cm} M{2.5cm} M{2.25cm} M{1.25cm}}
        \toprule
        Reference / Model & NAS Optimizer & Search Space & Eval. Strategy & \multicolumn{2}{c}{Continual Learning Type} & Benchmark & \makecell{Error \\ (avg. \%)} \\
        \midrule
        
         \makecell{Li et al. 2019 \\ (Learn-to-Grow) \cite{li_2019}} & GO & Continuous & OS & Task-Incremental & Prior-Focused \& Parameter Isolation & \makecell{ImageNet + \\ CIFAR100 + \\ + 8 others (avg)} & \makecell{30.16, \\ 20.41, \\ 22.32 } \\ 
         
         \makecell{Huang et al. 2019 \\ (CNAS) \cite{huang_2019}} & RL & Layer-Wise & WI & Class-Incremental & Episodic Rehearsal & \makecell{CIFAR100 (k=10)} & \makecell{~36.50} \\ 
         
         \makecell{Pasunuru \& Bansal 2019 \\ (CAS) \cite{pasunuru_2019}} & RL & Cell-Based & LFE & Task-Incremental & Prior-Focused & \makecell{MSR-VTT + \\ MSVD } & \makecell{55.50, \\ 38.30} \\ 
         
         \makecell{Zhang et al. 2020 \\ (E$^{2}$NAS) \cite{zhang_E2NAS_2020}} & GO & Continuous & OS & Task-Incremental & Prior-Focused & \makecell{CIFAR10 + \\ CIFAR100 + \\ ImageNet16} & \makecell{6.11, \\ 27.95, \\ 34.33} \\ 
         
         \makecell{Zhang et al. 2020 \\ (REC) \cite{zhang_rec_2020}} & RL & Layer-Wise & WI & Task-Incremental & Episodic Rehearsal \& Prior-Focused & \makecell{Permuted MNIST + \\ MNIST Variation + \\ CIFAR100} & \makecell{4.30, \\ 28.50, \\ 40.30} \\ 
         
         \makecell{Wu et al. 2020 \\ (Firefly) \cite{wu_2020}} & MO & Neuronal, Cell-Based & WI & Class-Incremental & Parameter Isolation & \makecell{CIFAR100 (k=20)} & \makecell{8.97} \\ 
         
         \makecell{Du et al. 2021 \\ (ENAS-S) \cite{du_2021}} & EA & Cell-Based & LFE & Class-Incremental & Parameter Isolation & \makecell{CIFAR10, \\ CIFAR100} & \makecell{~10.00, \\ 26.50} \\ 
         
         \makecell{Niu et al. 2021 \\ (AdaXpert) \cite{niu_2021}} & RL & Cell-Based & WI & Class-Incremental & Data-Focused & \makecell{ImageNet100, \\ ImageNet1000} & \makecell{19.26, \\ 18.87} \\ 
         
         \makecell{Zhang et al. 2021 \\ (NSAS) \cite{zhang_2021}} & RS & Cell-Based & OS & Task-Incremental & Parameter-Isolation & \makecell{CIFAR10 + \\ CIFAR100 + \\ ImageNet16} & \makecell{6.45, \\ 29.31, \\ 57.86} \\ 
         
         \makecell{Qin et al. 2021 \\ (BNS) \cite{qin_2021}} & RL & Layer-Wise & OS & Task-Incremental & Episodic Rehearsal & \makecell{MNIST, \\ CIFAR10, \\ CIFAR100} & \makecell{0.13, \\ 8.60, \\ 17.61} \\ 
         
         \makecell{Mundt et al. 2021 \\ (DP-NAS) \cite{mundt_2021}} & RL & Layer-Wise & LFE & Task-Incremental & Generative Rehearsal & \makecell{MNIST + \\ FashionMNIST, \\ MNIST + \\ CIFAR10} & \makecell{0.21, \\ 0.63, \\ 23.69, \\ 34.85} \\ 
         
         \makecell{Gao et al. 2022 \\ (CLEAS) \cite{gao_2022}} & RL & Neuronal & WI & Domain-Incremental, Class-Incremental & Parameter Isolation & \makecell{Permuted MNIST, \\ Rotated MNIST, \\ CIFAR100} & \makecell{3.70, \\ 3.00, \\ 33.10} \\ 

        \bottomrule
    \end{tabular}
    \bigskip
    \caption[Caption for CANNs Table]{Comparison of existing Continually-Adaptive models (CANNs)}
    \label{tab:pann_models}
    \vspace{-2.5em}
\end{table*}


More often than not, a dataset contains redundant features that negatively impact the convergence of the model during training. The function of any neural layer is to essentially map an input feature vector $\vec{x} = [x_{1}, x_{2}, ... , x_{n}] \in \mathbb{R}^{n}$ to an output vector $\vec{y} = [y_{1}, y_{2}, ... , y_{m}] \in \mathbb{R}^{m}$ with a nonlinearity, which can be modelled as:

\begin{equation}
\hat{y} = \sum_{i=1}^{n}{x_{i} W_{i} + b_{i}} \: ,
\label{eq:dense_model}
\end{equation}

\begin{equation}
\vec{y} = f(\hat{y})
\label{eq:ols_estimator}
\end{equation}

Where $W = [w_{1}^{\top}, w_{2}^{\top}, ... , w_{n}^{\top}] \in \mathbb{R}^{n \times m}$ is the parameter matrix of the network, $\vec{b} = [b_{1}, b_{2}, ... , b_{m}] \in \mathbb{R}^{m}$ is the bias vector, and $f(\cdot)$ is an element-wise nonlinearity (e.g. Tanh, Sigmoid, etc.). 

A training matrix $X = [\vec{x}, \vec{y}]$ with correlated features (i.e collinearity or multi-collinearity) is linearly dependent and thus $X$ is not a full rank matrix. Given the high-dimensional nature of deep neural networks, where $m$ and $n$ are often in the order of thousands \cite{huang_2017}, rank-deficiency significantly impacts the training time of the model \cite{lee_2021}. Feature Selection not only speeds up the convergence of the optimization process, it also enables the use of techniques like Low-Rank Approximation that can significantly reduce the memory-footprint of the model by representing the weight tensor as multiple smaller tensors through low-rank factorization \cite{sainath_2013}.

Due its non-trivial and creativity-driven nature, Feature Transformation and Construction is usually considered the most human-dependent Feature Engineering phase \cite{waring_2020,he_2021}. Constructing new features extends the dataset and its generalization by synthesizing new features from existing ones (as opposed to Data Augmentation, which adds new samples/rows rather than columns), whereas Feature Transformation affects the speed of the optimization process. Linear feature transformations, such as normalization and standardization, boost the convergence of the training process by scaling the raw data distribution and ensuring the features have similar magnitudes. Additionally, the traditional use of a single learning rate in ANNs upholds the importance of unifying the features' scale; in a Gradient Descent training scenario, a global learning rate applied to unnormalized data will most likely propel each gradient/dimension at proportionally different magnitudes. 


Similar to Feature Selection, Dimensionality Reduction (often called Feature Extraction) aims to separate redundant data from effective information. The primary distinction between the two processes is that Dimensionality Reduction creates new compressed mappings of the original features rather filter them as with Feature Selection. Although neural networks perform Dimensionality Reduction inherently (e.g. pooling or convolution layers extract features from localized regions), methods such as Principal Component Analysis (PCA) can scale a neural network's input and reduce the number of operations performed, thereby minimizing the computational requirements for training \cite{ren_2016}.

Feature Engineering can be regarded as a stochastic optimization problem and therefore a number of traditional optimization algorithms can be used to automate the process. The limited Automated Feature Engineering frameworks that have been proposed are mainly based on Genetic Algorithms \cite{de_2018,junior_2018} and Reinforcement Learning \cite{chen_fe_2019}, where both approaches performed impressively well compared to baselines (random- and manual/human-based). With the current lack of a unified benchmark, however, automated Data Preparation frameworks cannot be formally ranked or compared.


\subsubsection{Neural Architecture Generation and Optimization}

The crucial phase of generating a network's topology for a given task has been discussed in depth in Section 3. Nevertheless, the continuous maintenance of a neural network should not be interpreted as a repeated NAS process; while reiterating a NAS algorithm for each encountered task can indeed result in a fully-autonomous continuous pipeline, the computational cost and time required to repeatedly search for architectures from scratch will likely be unfeasible for most applications. To address the expected high resource demand for continual NAS frameworks, a few models have been recently proposed that incorporate CL techniques within the architecture search process (e.g. finding an optimal NAS cell that best fits all encountered tasks and facilitates few-shot learning for future tasks \cite{pasunuru_2019}).

Proposed by Gao et al. in 2022, Continual Learning with Efficient Architecture Search (CLEAS) dynamically expands the network architecture whilst preserving previous knowledge through parameter isolation \cite{gao_2022}. While this approach experiences no forgetting, it scales poorly; for every encountered task, the network has to expand with no means of compressing superfluous parameters. Similarly, ENAS-S \cite{du_2021} and Learn-to-Grow \cite{li_2019} also experience monotonic expansion without a memory-boundary, which makes them unsustainable for long-term real world deployment.

Moreover, Continual NAS (CNAS) \cite{huang_2019}, REC \cite{zhang_rec_2020}, and BNS \cite{qin_2021} all require a dedicated memory to store subsets of training data for replay purposes. While this memory-retention strategy offers great control over the retraining of the models as well as the NAS process, they are not capable of handling an infinite stream of input due to the unrestricted external memory requirement.

An alternative approach proposed by Nïu et al. (dubbed AdaXpert) dynamically adjusts the neural architecture using a secondary model called the Neural Architecture Adjuster (NAA), which aims to search for a new topology only when deemed necessary (i.e the distance between task boundaries is greater than a given threshold) \cite{niu_2021}. Tested incrementally on ImageNet1000, one of the most challenging vision benchmarks, this approach achieved state-of-the-art performance in just 2.5 to 7 GPU days, proving that combining NAS and CL optimization techniques is computationally feasible. However, in the experiment proposed for AdaXpert, the NAA uses Wasserstein distance for every input to compute how far the new task boundary is from the existing trained model's probability distribution, which shows an implicit and unmentioned assumption that the input samples must be grouped prior to training. Given a real world scenario where samples are interleaved, the NAA will be prompted to adjust the network for every increment, rendering it essentially futile.

\subsubsection{Hyperparameter Optimization}

Most NAS frameworks use the same set of hyperparameters over the search process \cite{he_2021}. While static hyperparameters might be viable for single-task NAS problems, they cannot be efficiently generalized for unknown future tasks, therefore continuous Hyperparameter Optimization is needed to maintain efficiency across sequentially encountered tasks.

Furthermore, most NAS frameworks consider the architecture and hyperparameter optimization processes as two separate phases \cite{zhang_rec_2020,he_2021}, which exponentially increases the complexity of the framework. To overcome nesting architecture search and hyperparameter tuning, jointly optimizing both the structure and its hyperparameters can significantly reduce the framework's requirements, especially on a continual basis. A number of studies on evolutionary ANNs suggest that hyperparameters can be endogenous features (i.e encoded as genes) and therefore get bootstrapped into the optimization process \cite{soltoggio_2018}.

\subsection{Desiderata for CANNs} \label{subsec:desiderata}

While the formal definition previously proposed in Section \ref{subsec:pann_definitions} describes the minimum qualification requirements for continual adaptiveness, there are several desirable qualities that could assist in generalizing CANNs towards a more general form of AI. The desiderata curated below stem from a holistic adaptiveness perspective, as well as an individualistic CL/NAS standpoint.

\begin{enumerate}
  \item \label{item:domain_agnostic} \textit{Domain-agnostic}: The model should not rely on a priori task-boundary definition; be capable of adapting to its surrounding environment without exhibiting a preference to a certain domain.
  
  \item \textit{Abstract representation}: Encoding the network should be represented with a high level of abstraction (i.e in the case of Evolutionary Computation, the genotype-phenotype relation should be indirect). This ensures model-scalability and follows on the biological analogy where the human genome is indirectly mapped to physical body \cite{stanley_2019}.
  
  \item \textit{Dynamic architecture with bounded capacity}: The network topology should grow, shrink, or get remapped to accommodate different input complexities and maintain efficiency. However, the architecture-search process should be bounded by a given memory capacity to avoid overflow and facilitate deployment in real-world applications.
  
  \item \label{item:graceful_forgetting} \textit{Graceful Forgetting}: With a bounded memory capacity and a potentially unbounded stream of data, forgetting is inevitable. It should, nevertheless, be imposed gracefully through selective means, where the less important tasks' performance gradually decays rather than catastrophically fails.
  
  \item \textit{Few-Shot Learning capability (positive FWT)}: Positive FWT is not often the focus for most Continual Learning models; however, for CANNs, facilitating Few-Shot Learning is a priority in order to reduce the large computational costs over time.
  
  \item \textit{Multi-Modal input support}: To further pave the way for a more generalized form of AI, a CANN framework should ideally support knowledge-acquisition through different modalities (e.g. image classification, text classification).
  
  \item \textit{Architecture Search Space inference}: The effectiveness of NAS frameworks is almost exclusively influenced by the Search Space design \cite{yu_2020}, which is typically a human-dependent process. Dynamic Search Space design for NAS is a favorable quality for CANNs to facilitate domain-agnostic adaptiveness.
  
  \item \textit{No explicit data storage for previous tasks}: Storing subsets of examples from previous tasks usually entails linearly increasing memory-consumption. Parameter-Constraint approaches, where the model's ``memory" is implicitly stored through the training process, are therefore considered a more favorable memory-retention technique. Additionally, the globally growing privacy concerns also dictate less data-storage-based methodologies.
  
  \item \textit{Learning rule plasticity}: The training process for ANNs greatly relies on predefined learning rules (or weight-updating equations) that determine the changes to the network parameters. Jointly optimizing the learning rule with the other network parameters (as suggested by \cite{soltoggio_2018,stanley_2019}) could effectively cut down the training cost.
  
\end{enumerate}

Although these features do not directly contribute to end-to-end automation and continuity, they aim to generalize CANNs and facilitate their scalability for real-world deployment.

\section{Future Directions} \label{sec:future_directions}


A palpable concern regarding CANNs is the potentially staggering computational requirements; pipelining multiple automated optimization approaches simultaneously with a possibly expanding network can be computationally expensive. Nevertheless, several models (Table \ref{tab:pann_models}) have showcased the feasibility of CANNs in all Incremental Learning settings. Moreover, as per criterion 2 (\ref{subsec:pann_definitions}-\ref{item:criterion_2}), Continual Learning should prospectively be incorporated within the NAS process to enable knowledge transfer across learning increments, further lowering the optimization costs. This fundamental inclusion of CL into the learning process has never been explored.

While AutoML frameworks (NAS specifically) and Continual Learning models have implicitly agreed upon benchmarks in the literature, the intersection between both fields does not (as shown in Table \ref{tab:pann_models}). Some works use the average accuracy over all tasks to rank their model \cite{li_2019,niu_2021}. However, that metric diminishes as the number of tasks approaches infinity. Consider a model with a bounded memory capacity for $\mathbb{C}$ tasks and an unbounded incremental input $x \:,\; s.t. \; |x| = +\infty$, the running average of accuracies will reach an inflection point at $x_{\mathbb{C}}$ where the inevitable forgetting will set in and the average value is bound to diminish. Furthermore, in the context of a system that enables an infinite stream of input, the model should not be ranked according to the ``final" model metrics; by definition, the framework has no final state and intermediate results might not reflect the framework's true performance. Additionally, the disjoint incremental learning settings and benchmarks cannot accurately estimate the models' adaptiveness and performance. Hence, a robust ranking approach for CANNs (potentially based on the rate of change of metrics rather than instantaneous values) needs to be formalized.


Furthermore, inconsistent with their state-of-the-art performance, Metaheurstic Optimization for NAS is often overlooked in comprehensive surveys and reviews \cite{elsken_2019,ren_2022}. Scarce coverage of such approaches limits the exploration of the optimization family, which has continued to show great potential in multiple areas and inherently fits the parallel exploration nature of NAS \cite{niu_2019,lankford_2020,mo_2021}. Some MO approaches, such as Microbial Intelligence and Glowworm Algorithm, have never been explored in the context of Neural Architecture Search, making them prospective future research directions.

Another gap in the NAS domain is the reliance on predefined search spaces. Yu et al. have concluded that the effectiveness of NAS frameworks is primarily attributed to how the Search Space is constrained \cite{yu_2020}; NAS models can achieve state-of-the-art performance with minimal cost if they are explicitly bound to a subset of the Search Space where optimal solutions exist. In the CANNs domain, however, a priori knowledge should not be exploited (Section \ref{subsec:desiderata}-\ref{item:domain_agnostic}), and thus domain-agnostic Search Space-inference methods are necessary.

A relevant research area that is adjacent to CANNs is \textit{Social Learning} (sometimes dubbed \textit{collaborative learning}). Human intelligence could not have evolved through one individual in a single life-time; similarly, it is unlikely that a complex set of mechanisms could evolve or be learned in a relatively short period of time without knowledge-sharing across multiple artificial models. Recent works on ML started exploring the concept of Social Learning, where a ML model would share information with other models in the same domain to collectively boost performances \cite{boyd_2011,soltoggio_2018}. Social Learning can be complementary to CANNs, especially given their lifelong learning nature.

Although unsupervised self-managing ANNs have been briefly explored in the literature \cite{mouret_2014,soltoggio_2018}, they are exclusively researched from an evolutionary/synaptic plasticity perspective. The unsupervised counterparts of NAS and CL (dubbed Unsupervised NAS (UnNAS) \cite{liu_unnas_2020} and Unsupervised Continual Learning (UCL) \cite{rao_2019}) have both shown promising potential individually, but their intersection has not been defined or reviewed.

One of the primary bottlenecks for full autonomy in CANNs is the need for manual definitions of task boundaries. The biological counterpart of lifelong learning is contingent on spontaneous generation of goals, which are fundamentally driven by intrinsic primal disposition to survival and procreation \cite{gopnik_1999}. Modelling these \textit{goal-seeking neural networks} \cite{portegys_2006} is an open challenge in AI that should be investigated further to enable true autonomy.

Notwithstanding the exciting prospect of generalized AI, there are drastic risks associated with such forms of continual automation. Unmonitored data collection introduces the possibility of accidental dataset poisoning, where unwanted samples are fed to the model. In certain sensitive domains (UAVs, medical applications, industrial IoT devices, etc.), data-integrity must be verified and input data should possibly be restricted to certain distribution boundaries as well. Safeguarding CANNs against potential task deviations would further pave the way for safe AI.

\section{Conclusion}

In this study, we propose a dynamic paradigm that entails full-automation and continual-adaptiveness in neural networks. Despite the existence of several continually-adaptive models (as per Table \ref{tab:pann_models}), this intersection between NAS and Continual Learning has never been formalized or reviewed in depth.

Through self-development and lifelong plasticity, CANNs aim to overcome numerous limitations posed by the human factor and result in more optimal models. In doing so, Continually-Adaptive Neural Networks could potentially introduce a new spectrum of applications where the functionality of the model is pliable, even after the deployment phase when remote access is often limited or unavailable.

Additionally, CANNs take a step towards generalizing intelligence; although the definition of ``intelligence" has long eluded psychologist and neuroscientists, it has been repeatedly defined as "goal-directed adaptive behavior" and ``adaptive self-modification" rather than mere associative inference \cite{sternberg_1982,goertzel_2014}. Narrow AI has indeed led to amazing accomplishments, however, there is an undeniable impending bottleneck to the capabilities that can be achieved with traditional methodologies. Continual adaptability and automation, on the other hand, minimize the static nature of Narrow AI and provide an essential basis for General AI. 

The proposed categorization schemes and analysis of methodologies aim to propel researches towards a less dedicated form of intelligence and facilitate more robust models that are sensitive to their surrounding environments.


\bibliographystyle{IEEEtran}
\bibliography{main.bib}

\end{document}